\documentclass[11pt,a4paper]{article}
\usepackage[hyperref]{conll-2019}
\usepackage{times}
\usepackage{latexsym}
\usepackage{subcaption}
\usepackage{array,multirow}
\usepackage{hyperref}       
\usepackage{url}            
\usepackage{booktabs}       
\usepackage{amsfonts}       
\usepackage{nicefrac}       
\usepackage{microtype}      
\usepackage{xcolor}
\usepackage{graphicx}
\usepackage{multicol}
\usepackage{amsmath}
\usepackage{url}
\usepackage{tabularx}\usepackage{graphicx,calc}
\newlength\myheight
\newlength\mydepth
\settototalheight\myheight{Xygp}
\usepackage{url}

\aclfinalcopy 

\title{Investigating Entity Knowledge in BERT with Simple Neural End-To-End Entity Linking}

\author{Samuel Broscheit \\
  Data and Web Science Group, University of Mannheim, Germany \\
  {\tt broscheit@informatik.uni-mannheim.de} \\}

\date{}
 
\begin{document}
\maketitle
\begin{abstract}
A typical architecture for end-to-end entity linking systems consists of three steps: mention detection, candidate generation and entity disambiguation. In this study we investigate the following questions: (a) Can all those steps be learned jointly with a model for contextualized text-representations, i.e. BERT  \citep{DBLP:conf/naacl/DevlinCLT19}? (b) How much entity knowledge is already contained in pretrained BERT? (c) Does additional entity knowledge improve BERT's performance in downstream tasks? To this end, we propose an extreme simplification of the entity linking setup that works surprisingly well: simply cast it as a per token classification over the entire entity vocabulary (over $700K$ classes in our case). We show on an entity linking benchmark that (i) this model improves the entity representations over plain BERT, (ii) that it outperforms entity linking architectures that optimize the tasks separately and (iii) that it only comes second to the current state-of-the-art that does mention detection and entity disambiguation jointly. Additionally, we investigate the usefulness of entity-aware token-representations in the text-understanding benchmark GLUE, as well as the question answering benchmarks SQUAD~V2 and SWAG and also the EN-DE WMT14 machine translation benchmark. To our surprise, we find that most of those benchmarks do not benefit from additional entity knowledge, except for a task with very small training data, the RTE task in GLUE, which improves by 2\%. 

\end{abstract}

\section{Introduction}

The goal of entity linking is, given a knowledge base (KB) and unstructured data, e.g. text, to detect mentions of the KB's entities in the unstructured data and link them to the correct KB entry. The entity linking task is typically implemented by the following steps: 
\begin{itemize}
    \item Mention detection (MD): text spans of potential entity mentions are identified,
    \item Candidate generation (CG): entity candidates for each mention are retrieved from the KB, 
    \item Entity disambiguation (ED): (typically) a mix of useful coreference and coherence features together with a classifier determine the entity link.
\end{itemize}
\citet{DBLP:journals/tacl/DurrettK14/JointEntity} were the first to propose jointly modelling MD, CG and ED in a graphical model and could show that each of those steps are interdependent and benefit from a joint objective. Other approaches only model MD and ED jointly \citep{DBLP:journals/tacl/NguyenTW16/JNERD, kolitsas-etal-2018-end/Ganea}, thus these architectures depend on a CG step after mention detection. \citet{Hachey:2013:EEL:2405838.2405914/CandGen, DBLP:conf/nldb/GuoQLLL13/CandGen, DBLP:journals/tacl/DurrettK14/JointEntity} showed the influence of CG on entity linking, because it can be the coverage bottleneck, when the correct entity is not contained in the candidates for ED. \citet{DBLP:conf/conll/YamadaS0T16, DBLP:journals/tacl/YamadaSTT17} use a precomputed set of entity candidates published by \citet{DBLP:conf/naacl/PershinaHG15} for their experiments on the CoNLL’03/AIDA benchmark dataset~\citep{DBLP:conf/emnlp/HoffartYBFPSTTW11/AIDA}, and due to this their experiments are comparable across studies with regards to the CG step. MD has a similar impact on entity linking performance, as it determines the upper bound of linkable mentions. 

BERT~\citep{DBLP:conf/naacl/DevlinCLT19} is a deep self-attention-based architecture which is pretrained on large amounts of data with a language modelling objective. This model provides very rich linguistic text-representations that have been shown to be very useful for many NLP tasks. Since its appearance, BERT is being analyzed and applied in various domains \citep{DBLP:journals/corr/abs-1903-10676/scibert, DBLP:journals/corr/abs-1901-08746/biobert}. A recent study found that BERT automatically learns the NLP pipeline \citep{DBLP:conf/acl/TenneyDP19}, i.e. a stack of increasingly higher level linguistic functions.
\citet{DBLP:journals/corr/abs-1905-07129/ernie} investigated injecting entity knowledge from noisy\footnote{TagMe's performance on various benchmark datasets ranges from 37\% to 72\%. F1 \citep{kolitsas-etal-2018-end/Ganea}} automatic entity linking into the pretraining of BERT and they could show that this improves relation extraction.

\begin{figure}
	\centering
	\includegraphics[scale=0.9]{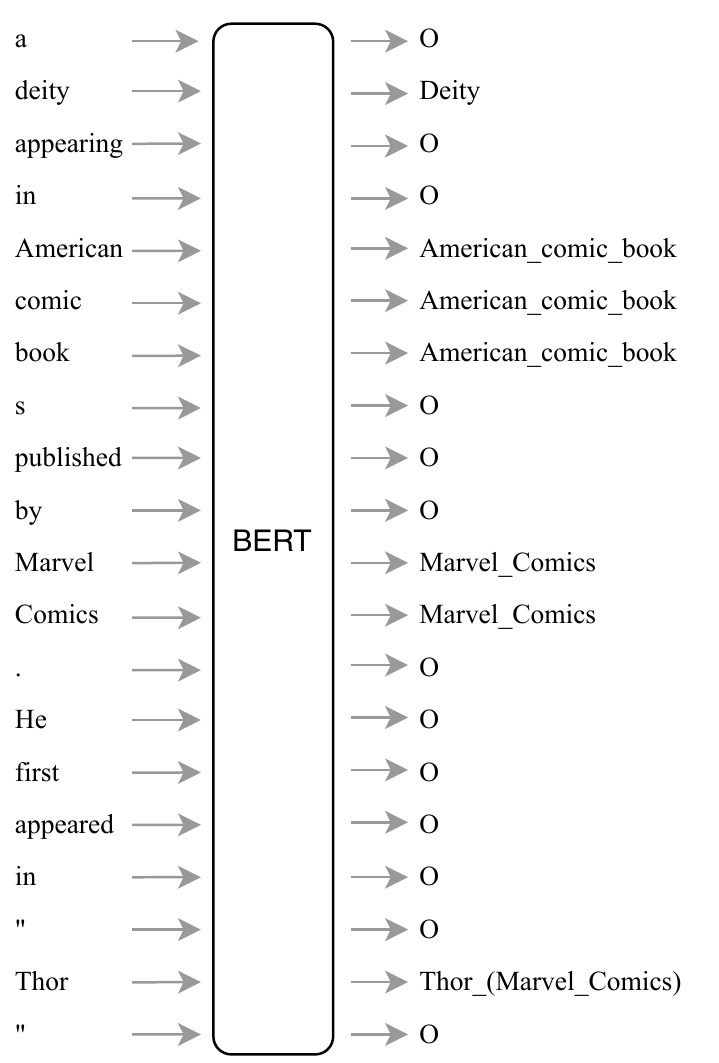}
	\caption{Illustrating the simple neural end-to-end entity linking setup. BERT+Entity predicts entity links per token, where "O" denotes a \textit{Nil} prediction. The example shows how context can help to link \textit{``Thor''} to \textit{Thor\_(Marvel\_Comics)}.} 
	\label{fig:end2endel}
\end{figure}

In this study we investigate the following questions:  

\paragraph{(a) Can BERT's architecture learn all entity linking steps jointly?} We propose an extreme simplification of entity linking and cast it as a per token classification over the entire entity vocabulary, thus solving MD, CG and ED simultaneously (see Fig. \ref{fig:end2endel}). The entity vocabulary is based on the $700K$ top most frequent entities in English Wikipedia and the training data was derived from English Wikipedia texts. We first trained BERT-base-uncased on English Wikipedia (dubbed BERT+Entity) and then fine-tuned and evaluated it on an entity linking benchmark. We found that this worked surprisingly well for entity linking, even if we do not have any supervision on mention-spans, i.e. BIO tags. An error analysis with validation data revealed that only 3\% of errors are purely due to span errors, while most errors are due to wrong \textit{Nil} predictions which often coincided with entities being infrequent.

\paragraph{(b) How much entity knowledge is already contained in pretrained BERT?} To investigate this question, we froze BERT and only trained the entity classifier of BERT+Entity on Wikipedia (dubbed Frozen-BERT+Entity), i.e. the resulting entity classifier is adjusted for entity mentions for which plain BERT already does assign distinct token representations, such that correct entity classification is possible. Then we fine-tuned and evaluated Frozen-BERT+Entity on an entity linking benchmark. We find that the performance of Frozen-BERT+Entity is 6\% below BERT+Entity, showing that BERT+Entity has learned additional entity knowledge. 

\paragraph{(c) Does additional entity knowledge improve BERT's performance in downstream tasks?}
Due to training BERT+Entity with a per token classification, the model is forced to assign distinct entity specific features to each token of an entity mention. Downstream tasks could exploit this, if additional entity information is necessary for them. We evaluated BERT+Entity in the natural language understanding benchmark GLUE~\citep{DBLP:conf/emnlp/WangSMHLB18/GLUE}, the question answering (QA) benchmarks SQUAD~V2 \citep{DBLP:conf/acl/RajpurkarJL18/SQUADV2} and SWAG \citep{DBLP:conf/emnlp/ZellersBSC18/SWAG}, and the machine translation benchmark EN-DE~WMT14. We confirm the finding from  \citet{DBLP:journals/corr/abs-1905-07129/ernie} that additional entity knowledge is not beneficial for the GLUE benchmark. To our surprise, we also find that additional entity knowledge is neither helpful for the two QA datasets nor for machine translation. The only exception is the RTE task in GLUE in which BERT+Entity improves 2\%. This dataset has just 0.5-2\% of the training data of the two larger natural language inference datasets in GLUE.    

Our contributions are: We are the first to study the latter questions. We are also the first to propose a fully neural model, that does MD, CG and ED all in one model, i.e. performing entity linking without any pipeline or any heuristics. We are also the first to propose to model entity linking as a token classification and show that this seems to be a viable option. We also uncover that there is a lack of tasks that evaluate additional entity knowledge in pretrained language models.

\section{Related Work}

\paragraph{Entity Linking} \citet{DBLP:journals/tacl/DurrettK14/JointEntity} is the work that is closest to our approach, although not neural. In their approach they model interactions between the MD, CG and ED tasks jointly. They find that the joint objective is beneficial, such that each task improves. They also note that there is no natural order of the tasks and they should interact freely. Their approach to CG is to learn to generate queries to the KB. \citet{DBLP:journals/tacl/NguyenTW16/JNERD} also propose jointly modelling MD and ED  with a graphical model and show that it improves ED performance and is more robust. \citet{kolitsas-etal-2018-end/Ganea} recently published their study in which they propose the first neural model to learn MD and ED jointly. Their proposed method is to overgenerate mentions and prune them with a mention-entity dictionary. The ED step reasons over the remaining mentions if and to what they link to. However, modern approaches for solving natural language tasks operate on neural text-representations, and the approaches discussed so far only yield entity-links. \citet{DBLP:conf/conll/YamadaS0T16, DBLP:journals/tacl/YamadaSTT17} was the first to investigate neural text representations and entity linking, but their approach is limited to ED.

\paragraph{Pretrained Language Models} ULMFIT~\citep{DBLP:conf/acl/RuderH18}, ELMO~\cite{DBLP:conf/naacl/PetersNIGCLZ18}, BERT~\citep{DBLP:conf/naacl/DevlinCLT19} and GPT-2 \citep{radford2019language} are modern language models that are very deep and wide (for NLP) and are pretrained on large amounts of data. They provide very rich text representations that have shown to improve many NLP tasks by just replacing the static word embeddings with deep contextualized word embeddings. As  \citet{DBLP:conf/rep4nlp/PetersRS19} show, further training the deep language models alongside the model that uses the embeddings as input can be helpful, for which the term ``finetuning'' is used. The current trend in research is to investigate all aspects of these language models, seeking insights in their inner workings \citep{DBLP:conf/acl/TenneyDP19}, or their application to various domains \citep{DBLP:journals/corr/abs-1903-10676/scibert, DBLP:journals/corr/abs-1901-08746/biobert}. In this study, we investigate the factual information in form of entities that is contained in BERT, seeking to understand to what degree this information is already identifiable in BERT and if the entity knowledge can be improved.

\section{End-To-End Neural Entity-Linking}

In this section we describe the BERT+Entity, which is a is straightforward extension of BERT, however, as with the original BERT, the main challenge lies in designing the training scheme, i.e. in our case the creation of the training data. 
Our goal for the experiments is to evaluate, if we can learn candidate generation, thus a desiderata is to make the entity vocabulary as large as possible to be comparable to other studies. The text data and the entity linking annotations are derived from Wikipedia by exploiting intra-Wikipedia links. This yields the challenge that the annotations for entity links from Wikipedia are assumed to be incomplete, i.e. not every entity mention in Wikipedia is linked, which we hypothesize can be detrimental during training.  

\subsection{Model}
Our model is based on BERT, which is a deep self-attention-based architecture \cite{DBLP:conf/nips/VaswaniSPUJGKP17/AttentionTransformer} that was trained on large amounts of text. Its training objective is two-fold: (a.) predict missing tokens from sentences, and (b.) classify if a second sentence was an adjacent sentence. The input and output token vocabulary are sub-words, i.e. the vocabulary is computed from the training data by determining the $30K$  most frequent character sequences, excluding spaces. \citet{DBLP:conf/naacl/DevlinCLT19} made several pretrained BERT models publicly available. They differ in size --- i.e. token embedding size and self-attention layer depth --- and whether the token vocabulary is cased or uncased. BERT+Entity is a straightforward extension on top of BERT, i.e. we initialize BERT with the publicly available weights from the BERT-base-uncased model and add an output classification layer on top of the architecture. Given a contextualized token, the classifier computes the probability of an entity link for each entry in the entity vocabulary. Formally, let $d$ be BERT's token embedding size, and $E \in \mathbb{R}^{|KB|\times d}$ the entity classification layer, with $|KB|$ being the number of entities in the KB, $V$ is the sub-word vocabulary, $c_i = BERT(h)[i]$ is the $i\text{-th}$ contextualized token computed by BERT from context $h = [v_1, v_2, ..., v_{i-1}, v_{i}, v_{i+1}, ..., v_{m} ]$ with each $v \in V$. Consequently, the probability $p(j|v, h)$ of word $v$ --- which is the $i\text{-th}$ token in context $h$ --- linking to entity $j$ is computed by $\sigma(E_jc_i)$, where $\sigma$ is the sigmoid function.

\subsection{Training Data}
\label{sec:training_data}

The entity vocabulary and training data are derived from English Wikipedia texts\footnote{From a enwiki Wikipedia dump from 20.06.2017.}. We used an extended version of WikiExtractor\footnote{https://github.com/samuelbroscheit/wikiextractor-wikimentions} to extract the text spans that are associated with an internal Wikipedia link to use as annotation, e.g. in the sentence \textit{``The first \underline{Thor} was all about introducing Asgard''}, the text span \textit{``Thor''} links to \textit{https://en.wikipedia.org/wiki/Thor\_(film)}. BERT is originally trained with sentences. However, for entity linking, a larger context can help to disambiguate entity mentions, which is why we select text fragments of such a length, that they span multiple sentences. For later use we collect $(m,e)$ tuples of entities $e$ and their mention $m$. This yields a set $M$ of potentially linkable strings and also lets us compute the conditional probability $p(e|m)$ based on the $\#(m,e)$ counts.  

\paragraph{Handling incomplete annotation} A challenge in using the Wikipedia links as annotation is that most entities do not have all their mentions annotated, i.e. often only the first appearance in an article is linked. We hypothesize that learning a classifier on such skewed data would yield a skewed model. Our approach to counter missing annotations is two-fold: (i) We only select text fragments that contain a minimum count of annotated Wikipedia links. (ii) To account for unlinked mentions in the fragments we use a Trie-based matcher\footnote{https://github.com/vi3k6i5/flashtext} to annotate all occurrences of  linkable strings that we collected in $M$. As entity links we annotate all possible entities this mention could link to but only with the conditional probability $p(e|m)$, with the goal that the model remembers a context independent entity prior. One issue is that due to the incomplete annotation, the $\#(e,m)$ counts yield $p(Nil|\textit{``United States''})>0$, i.e. the mention \textit{``United States''} has a large non-zero probability to link to nothing. Based on the assumption that the mentions of the most popular entities should always link to something, we compute the average of the probability of linking to \textit{Nil} for the $k=1000$ most frequent entities 
\[
\Bar{p}_{Nil}=\frac{1}{k}\sum_j \frac{\#(m_j,Nil)}{\#m_i}.
\] 
and use $\#(m_i,Nil) - \frac{\Bar{p}_{Nil}}{(1-\Bar{p}_{Nil})}*\#(m_i,e_*)$ to discount $\#(m_i,Nil)$ such that $p(Nil|\textit{``United States''})\approx0$, i.e. the model should always link ``United States'' and mentions of less frequent entities get an increase in probability to link to something. 

\section{Entity Linking Experiments}

In the experiments we want to investigate how the simple neural end-to-end entity linking model BERT+Entity performs, i.e. if it learns something additional on-top of BERT. Additionally, we investigated if the entity-aware token-representations are useful for downstream tasks. We also discuss the main engineering challenges training with such a large entity vocabulary.  

\subsection{Data}

\paragraph{Wikipedia}
\label{sec:data}
We report two settings which differ in size of the entity vocabulary, size of the fragments and minimum number of entities per fragments. The first setting was the initial study, and the second one is a follow up study in which we changed settings that potentially could improve entity linking performance.

\textbf{Setting I}: We keep the $700K$ top most frequent entities from the $\approx6M$ entities in Wikipedia, i.e. we chose the entity vocabulary as large as it was technically feasible with regards to memory and training speed. To put it into context, the CoNLL’03/AIDA entity linking benchmark contains $23,5K$ entities in $1300$ documents. We are missing $30$ entities from CoNLL’03/AIDA that only appear less than $10$ times in the Wikipedia training data. We chunk the Wikipedia texts into fragments with a length of $110$ tokens and an overlap of $20$ tokens with the previous and following fragment. We only keep fragments that contain at least $1$ infrequent linked entity or at least $3$ frequent ones. This yields $8,8M$ training instances from which we take $1000$ each for validation and testing. 

\textbf{Setting II}: We keep the $500K$ top most frequent entities, which is comparable to the entity vocabulary of \citet{kolitsas-etal-2018-end/Ganea} and we have to add $\approx1000$ entities from CoNLL’03/AIDA to the entity vocabulary to be able to evaluate our model on that benchmark. We increase the fragment size to $250$ tokens and keep fragments that contain at least $1$ linked entity but keep at most $500$ fragments per entity. This yields $2,4M$ training instances from which we take $500$ each for validation and testing.

\paragraph{Entity Linking Benchmark}

To evaluate on a commonly used benchmark dataset we use CoNLL’03/AIDA. It is the biggest manually annotated ED dataset. It contains 946 documents in training, 216 in validation (testa/AIDA-VALID) and 231 in test (testb/AIDA-TEST).

\subsection{Training}

\begin{figure}
	\centering
	\includegraphics[scale=0.5]{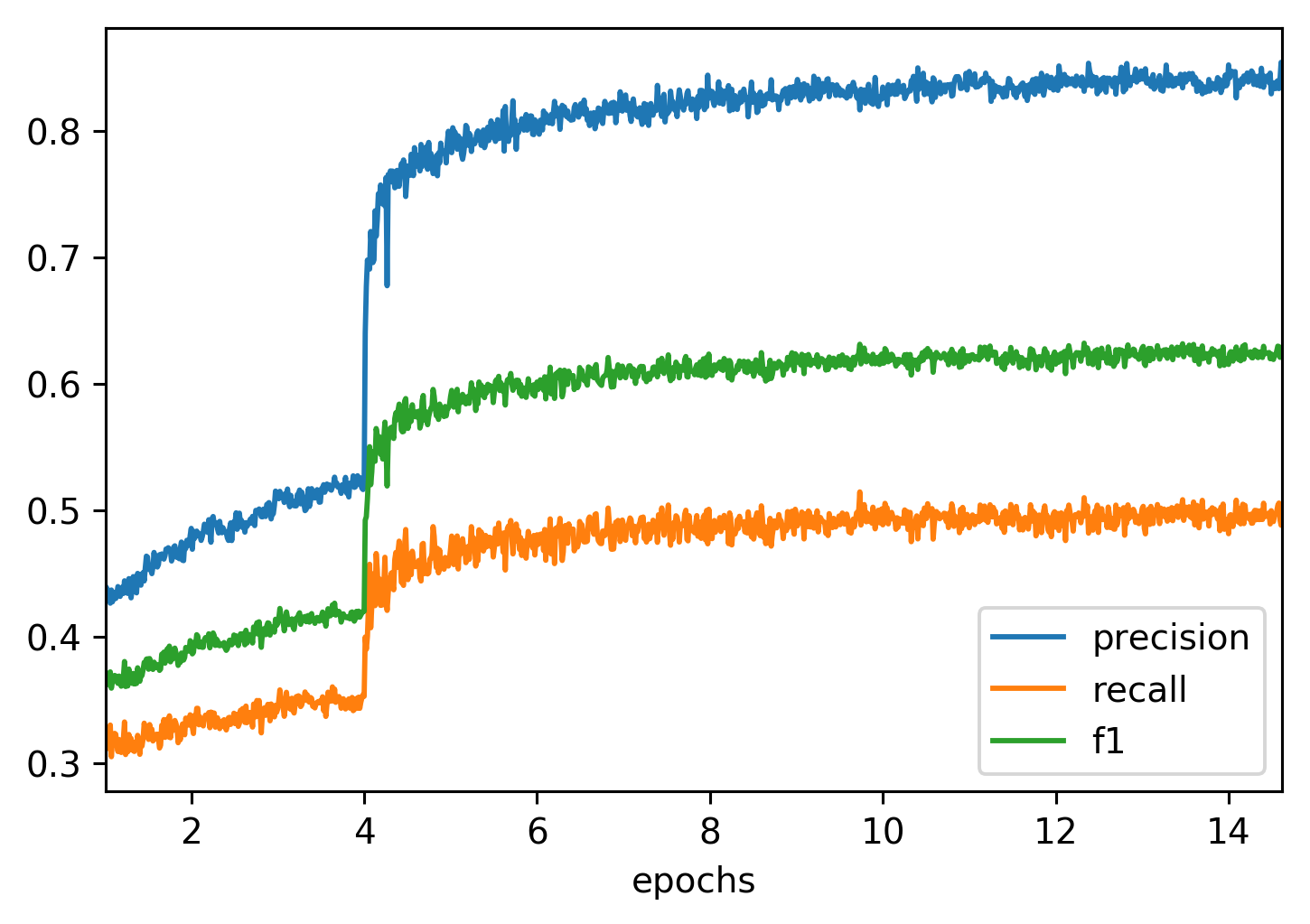}
	\caption{Per token classification InKB scores on the validation data during training on the Wikipedia dataset in Setting II for 40 days. The jump at the $4\text{-th}$ epoch happens when we switch from training Frozen-BERT+Entity to BERT+Entity, i.e. when we start fine-tuning BERT.}
	\label{fig:training}
\end{figure}

\begin{table*}[h]
\centering
\begin{tabular}{@{}llcclccl@{}}
\toprule
 & & \multicolumn{3}{c}{AIDA/testa}   & \multicolumn{3}{c}{AIDA/testb}                               \\ 
 & & \multicolumn{1}{l}{strong F1}    & \multicolumn{1}{l}{weak F1} & ED   & \multicolumn{1}{l}{strong F1} & \multicolumn{1}{l}{weak F1} & ED  \\ \midrule \midrule
\multicolumn{2}{@{}l}{\citet{kolitsas-etal-2018-end/Ganea} indep. baseline} & 80.3     & 80.5     & -        & 74.6     & 75.0     & -        \\
\multicolumn{2}{@{}l}{\citet{kolitsas-etal-2018-end/Ganea}}                 & 89.4     & 89.8     & 93.7     & 82.4     & 82.8     & 87.3     \\ \midrule
 BERT &                                                                  & 63.3     & 66.6     & 67.6     & 49.6     & 52.4     & 52.8     \\ \midrule
 Setting I & Frozen-BERT+Entity                                          & 76.8     & 79.6     & 80.6     & 64.7     & 68.0     & 68.6     \\
           & BERT+Entity                                                 & 82.8     & 84.4     & 86.6     & 74.8     & 76.5     & 78.8     \\ \midrule
 Setting II& Frozen-BERT+Entity                                          & 76.5     & 80.1     & 79.6     & 67.8     & 71.9     & 67.8     \\
           & BERT+Entity                                                 & 86.0     & 87.3     & 92.3     & 79.3     & 81.1     & 87.9     \\ \bottomrule \bottomrule
\end{tabular}
\caption{Comparing entity linking results on CoNLL’03/AIDA. \textit{strong F1} and \textit{weak F1} denote InKB F1 scores. \textit{ED} is Precision@1 for InKB. \citet{kolitsas-etal-2018-end/Ganea} also study a neural model, however, they only model MD and ED. The independent baseline shows how their model performs when they use mentions detected by Stanford NLP. In Frozen-BERT+Entity BERT is not trained and only the entity classifier on-top is trained. }
\label{tab:resultsel}
\end{table*}

We use a multi-class classification over the entity vocabulary, i.e. the label $y$ vector for one token $v_i$ is defined by
\begin{align*}
    y_{ij}= p(j|v_i),\  
\text{for } j \in \{1, .., ||KB||\}.
\end{align*}

\noindent However, computing the loss over the whole entity vocabulary would be infeasible, because the entity mention vocabulary is very large and the gradients for the entity classifier would exceed our GPU memory. Thus, to improve memory efficiency and increase convergence speed, we use negative sampling. After sampling text fragments for a batch $b$, we collected the set $N+_b$ of all true entities ---according to the annotations discussed in Sec. \ref{sec:training_data} ---  that occurred in those text fragments. Ideally we would update the representations of those entities that do not occur in the set $N+_b$ which the model is erroneously the most confident about. To achieve this, we first performed a prediction for the text fragments in the current batch and collected for each token the top $k$ predicted entities. We aggregated the entities' logits over the whole batch and sorted the entities by their aggregated logits into the list $N_b-$ and removed from it any entity contained in $N_b+$. We join $N_b = {N_b+} \cup N_b-$ and truncate $N_b-$ such that $|N_b|$ equals a given maximum size. Each label vector $y_i$ for token $c_i$ from fragment $C$ in batch $b$ was now defined over the entities in $N_b$. Thus, we only predict over the corresponding subset of the entity embedding table, i.e.  $\hat{E} = E(N_b)$. The loss for one fragment $C$ in batch $b$ was computed by 
\begin{align*}
    L = \frac{1}{|N_b|*|C|}\sum_{i}^{|C|}\sum_{j}^{|N_b|} - [ y_{ij} \cdot \log \sigma(\hat{E}_{j}c_i) \\
        + (1 - y_{ij}) \cdot \log (1 - \sigma(\hat{E}_{j}c_i))].
\end{align*}

For training on Wikipedia we used Adam \citep{DBLP:journals/corr/KingmaB14} with mini batch size 10, gradient accumulation over $4$ batches, maximum label size $10240$, the learning rate for BERT was $5e\text{-}5$ and for the entity classifier $0.01$. In Setting I we train the model for 4 epochs, one epoch took five days with two TitanXp/1080Ti. In the first 1.5 epochs we train Frozen-BERT+Entity and then BERT+Entity. In Setting II we train the model for 14 epochs and one epoch took three days. In the first 3 epochs we train Frozen-BERT+Entity and then BERT+Entity.  

For training on CoNLL’03/AIDA we used Adam \citep{DBLP:journals/corr/KingmaB14} with mini batch size 10, gradient accumulation over $4$ batches, maximum label size $1024$, learning rates for BERT $5e\text{-}5$, dropout in BERT $0.2$, and we freeze the token embeddings, the first two layers of BERT and the entity classifier. We train the remaining parameters for up to 30 epochs and perform early stopping according to strong match (see next Section). One epoch took seven minutes with one TITAN~Xp/1080~Ti.

\subsection{Performance Metrics}

We compute the Micro InKB Precision, Recall and F1 metrics and we only consider entities as true, if they are in our KB. We compute a strong match, i.e. every token in the gold annotated span has to be classified correctly. We also report a weak match, which we define as at least one token in the gold annotated span having to link to the correct entity. This setting accounts for annotation inconsistencies, e.g. when the model and the annotation do not agree on which mention \textit{``U.S. army''} or \textit{``U.S.''} to annotate (can be either way). We also report strong ED Precision@1, i.e. we ignore \textit{Nil} predictions of the model and only evaluate the top ranked entity only for spans that have a gold entity. 

\subsubsection{Results}

In Table \ref{tab:resultsel} we compare our results to the most recent results by \citet{kolitsas-etal-2018-end/Ganea} who studied a neural approach that does joint modelling of MD and ED, but not CG. They also provide a baseline in which they show how their classifier performs when MD and ED are independent, i.e. linking mentions detected by Stanford NLP. 

For the reported results denoted only with BERT, the entity classifier is trained from scratch on CoNLL’03/AIDA and BERT is finetuned. This shows the lower bound on this dataset, i.e. the amount of information that we can learn with BERT only from the CoNLL’03/AIDA training data. Note, that this cannot generalize to entities that are not contained in training. The difference between BERT and Frozen-BERT+Entity shows the amount of entity knowledge that plain BERT already had, which it transferred in the entity classifier during training on Wikipedia. Finally, BERT+Entity is the proposed model, in which both BERT and the entity classifier have been trained on Wikipedia.

\begin{table}[]
\centering
\begin{tabular}{@{}lr@{}}
\toprule
Reason for error               & \# \\ \midrule
no prediction                  & 57 \\
different than gold annotation &    \\
\ \ \ \ no obvious reason                & 13 \\
\ \ \ \ semantic close            & 4  \\
\ \ \ \ lexical overlap          & 5  \\
\ \ \ \ nested entity                  & 5  \\
gold annotation wrong          & 12 \\
span error                     & 3  \\
unclear                        & 1  \\ \bottomrule
                               & 100  \\ \bottomrule
\end{tabular}
\caption{Investigating the types of strong precision errors of BERT+Entity trained in Setting I on CoNLL’03/AIDA (testa) on 100 randomly sampled strong precision errors from the validation dataset.}
\label{tab:errorana}
\end{table}

\subsubsection{Discussion}

Comparing BERT+Entity and Frozen-BERT+Entity we see that there is a significant amount of entity knowledge that BERT+Entity learns additionally to Frozen-BERT+Entity, i.e. training BERT+Entity increases the scores between 6\%-10\% depending on the score and dataset. However, it should also be noted that Frozen-BERT+Entity already shows an increase of 13\%-16\% over BERT, thus it already learns for many entities distinct features that enable the entity classifier to identify them. The improvement of Frozen-BERT+Entity in contrast to BERT on CoNLL’03/AIDA shows that this pretraining generalizes to validation and test data. We can also observe that Setting II improves by a large margin over Setting I and comes very close to the results of \citet{kolitsas-etal-2018-end/Ganea}. We conjecture that the biggest impact on the performance from changing the training from  Setting I to Setting II, was due to the downsampling of the training data in favor of less frequent entities. This reduction of training data in Setting II --- caused by capping the maximum amount of examples per entity --- enabled us to run more epochs in less time, which might have improved the representations of less frequent entities.  

\begin{table*}[]
\centering
\begin{tabular}{@{}llll@{}}
\toprule
Task   & Metric                       & BERT-BERT-Ensemble    & BERT+Entity-Ensemble \\ \midrule
CoLA   & Matthew's corr.              & 59.92       & 59.97           \\
SST-2  & accuracy                     & 92.73       & 92.43           \\
MRPC   & F1/accuracy                  & 89.16       & 90.13           \\
STS-B  & Pearson/Spearman corr.       & 89.90       & 89.60           \\
QQP    & accuracy                  & 91.64       & 91.21           \\
MNLI   & matched acc./mismatched acc. & 84.96       & 84.78           \\
QNLI   & accuracy                     & 91.21       & 91.15           \\
RTE    & accuracy                     & 71.48       & 73.64           \\
WNLI   & accuracy                     & 56.33       & 56.33           \\ \midrule
SQUAD V2 & matched/mismatched           & 76.89/73.83 & 76.36/73.46     \\ \midrule
SWAG   & accurracy                    & 80.70       & 80.76           \\ \midrule
WMT14 EN-DE   & BLEU                        & 22.51       & 22.20           \\ \bottomrule
\end{tabular}
\caption{Experiments on downstream tasks with BERT+Entity trained in Setting I. The first group are the GLUE tasks, then followed by SQUAD V2 and SWAG (for which only the dev set results are reported), and the results for machine translation WMT14 EN-DE.}
\label{tab:downstream}
\end{table*}

When we compare BERT+Entity with the two results from \citet{kolitsas-etal-2018-end/Ganea}, we observe that BERT+Entity improves over the baseline that models MD, CG and ED independently, and that BERT+Entity comes second to the current state-of-the-art in end-to-end entity linking. What can also be observed is that the performance of all models drops from AIDA/testa to AIDA/testb. For BERT+Entity, however, the drop is more severe, obviously the model overfits to some patterns in the training data that are present in the validation data, but not in the test data. We hypothesize that this might be due to some sport specific documents that make roughly $1/4$ of the dataset's mentions. However, without spoiling the test-set we cannot know for sure. 

In Table \ref{tab:errorana} we performed an error analysis for the experiments for Setting I to learn what kind of strong precision errors are responsible for the performance of BERT+Entity. The largest source of errors was that BERT+Entity did predict \textit{Nil} instead of an entity. We hypothesized that most of the \textit{no prediction} errors are because those entities have only a low frequency in the training data, i.e. this could be solved by increasing the model size and improving the training time. Another source of error we observed was that the context size was too small due to the fragment size. A surprisingly positive result from the error analysis was that in only 3\% a wrong span caused the error. Motivated by the observations we devised the follow-up experiment Setting II (see Section \ref{sec:data}) in which we changed some of the settings to potentially solve the observed issues.

\section{Downstream Tasks Experiments}

In this section we discuss the downstream task results. We performed evaluations on the natural language understand task GLUE, the question answering tasks SQUAD V2 and SWAG and the machine translation benchmark EN-DE WMT14. We found that only in one of the subtasks of GLUE ---the natural language inference tasks RTE--- BERT+Entity performs better than BERT, for all other we can observe no such effect. The reported results are for Setting I, however, we repeated the experiments with Setting II and observed the same outcomes. 

\subsection{Model}

For the tasks GLUE, SQUAD V2 and SWAG we extend hugginface's implementation\footnote{https://github.com/huggingface/pytorch-pretrained-BERT} and concatenate the outputs of BERT and BERT+Entity (dubbed BERT+Entity-Ensemble) or two BERTs (dubbed BERT-BERT-Ensemble). For EN-DE WMT14 we use BERT (dubbed BERT-2Seq) or BERT+Entity (dubbed BERT+Entity-2Seq) as encoder and use a Transformer decoder by adapting fairseqs Pytorch Seq2Seq Transformer implementation \citep{ott2019fairseq}. 

\subsection{Training}

For the GLUE benchmark, SQUAD and SWAG we train the BERT+Entity-Ensemble and BERT-BERT-Ensemble for 3 epochs and use the default hyperparameters from the implementation. The models BERT-2Seq and BERT+Entity-2Seq we train for 4 epochs, with Adam as optimizer and learning rate $5e\text{-}5$, max $1000$ tokens per batch, clip gradient norm $0.1$, dropout $0.2$, label smoothing $0.1$, and we keep the encoders BERT and BERT+Entity fixed for the first epoch and then train it together with the decoder.  

\subsection{Results}
We find that the additional entity knowledge is not helpful in the evaluated tasks. The results in Table \ref{tab:downstream} show that, except for RTE, there seems to be no advantage in having additional entity knowledge. The question is, if this is (a) due to the entity overlap in training and testing such that also an entity unaware model can learn the necessary model, or (b) the entities are too scarce in the training data to make a difference, or (c) the tasks themselves do not require entity knowledge, i.e. other textual cues are enough. We leave those questions for future research.  

\section{Conclusion}

In this study we investigated an extremely simplified approach to entity linking that worked surprisingly well and allowed us to investigate entity knowledge in BERT. Even when there is a gap to the current state-of-the-art in entity linking, we hypothesize that this gap can be closed with larger hardware capacity to scale up the model size and effective training time. Apart from that, the model is the first that performs entity linking without any pipeline or any heuristics, compared to all prior approaches. We found that with our approach we can learn additional entity knowledge in BERT that helps in entity linking. However, we also found that almost none of the downstream tasks really required entity knowledge, which is an interesting observation and an open question for future research.  

\section*{Acknowledgments}

The author would like to gratefully thank the NVIDIA corporation for the donation of a TITAN Xp GPU that was used in this research.

\bibliography{conll-2019}
\bibliographystyle{acl_natbib}

\appendix

\end{document}